\documentclass[sigconf, nonacm]{acmart}

\usepackage{multirow}
\usepackage{graphicx}
\usepackage{lscape}
\usepackage{pifont}
\usepackage{subcaption}
\usepackage{hyperref}

\AtBeginDocument{%
  }

\begin{document}

\title{Dynamic Self-adaptive Multiscale Distillation from Pre-trained Multimodal Large Model for Efficient Cross-modal Representation Learning}

\author{Zhengyang Liang}
\email{1ce@bupt.edu.cn}
\affiliation{%
  \institution{Beijing University of Posts and Telecommunications}
  \city{Beijing}
  \country{China}
}

\author{Meiyu Liang}
\authornote{Corresponding Author}
\email{meiyu1210@bupt.edu.cn}
\affiliation{%
  \institution{Beijing University of Posts and Telecommunications}
  \city{Beijing}
  \country{China}
}

\author{Wei Huang}
\email{huangweichaojibang@bupt.edu.cn}
\affiliation{%
  \institution{Beijing University of Posts and Telecommunications}
  \city{Beijing}
  \country{China}
}

\author{Yawen Li}
\email{warmly0716@126.com}
\affiliation{%
  \institution{Beijing University of Posts and Telecommunications}
  \city{Beijing}
  \country{China}
}

\author{Zhe Xue}
\email{xuezhe@bupt.edu.cn}
\affiliation{%
  \institution{Beijing University of Posts and Telecommunications}
  \city{Beijing}
  \country{China}
}

\begin{abstract}
In recent years, pre-trained multimodal large models have attracted widespread attention due to their outstanding performance in various multimodal applications. Nonetheless, the extensive computational resources and vast datasets required for their training present significant hurdles for deployment in environments with limited computational resources. To address this challenge, we propose a novel dynamic self-adaptive multiscale distillation from pre-trained multimodal large model for efficient cross-modal representation learning for the first time. Unlike existing distillation methods, our strategy employs a multiscale perspective, enabling the extraction and distillation of structural knowledge across different dimensions from the pre-trained multimodal large model. Ensuring that the student model inherits a comprehensive and nuanced understanding of the teacher knowledge. Our distillation framework can capture important structural knowledge of the teacher's network and focus on hard negative samples. To optimize each distillation loss in a balanced and efficient manner, we propose a dynamic self-adaptive distillation loss balancer, a novel component eliminating the need for manual loss weight adjustments and dynamically balances each loss item during the distillation process. Our methodology streamlines pre-trained multimodal large models using only their output features and original image-level information, requiring minimal computational resources. This efficient approach is suited for various applications and allows the deployment of advanced multimodal technologies even in resource-limited settings. Extensive experiments has demonstrated that our method maintains high performance while significantly reducing model complexity and training costs. Moreover, our distilled student model utilizes only image-level information to achieve state-of-the-art performance on cross-modal retrieval tasks, surpassing previous methods that relied on region-level information.  Code is available at \href{https://github.com/chrisx599/DSMD}{https://github.com/chrisx599/DSMD}.
\end{abstract}

\renewcommand\footnotetextcopyrightpermission[1]{}
\settopmatter{printacmref=false}
\maketitle

\section{Introduction}
In recent years, pre-trained multimodal large models utilizing large 
computational resources and rich data have performed well on a wide range of downstream tasks and can be easily adapted for applications. CLIP\cite{CLIP} and ALIGN\cite{ALIGN} are recently proposed pre-trained multimodal large models based on contrastive learning, which form a task-independent model by predicting which text matches which image. In ALBEF\cite{ALBEF}, it applies contrastive loss to align image and text features before modeling their joint represen-tation. TCL\cite{TCL} uses intramodal and intermodal contrastive learning to guide the learning process through knowledge distillation. FACLC-L\cite{FACLCL} encodes structured event knowledge to augment visual-language pre-training with textual event maps for contrastive learning. BEiT-3\cite{BEiT-3} models images as foreign languages, modeling the masking "language" of images, text, and image-text pairs in a unified way, and aligning image and text features with contrastive learning.

However, these pre-trained multimodal large models require a lot of computational resources and are unable to be applied in resource-constrained scenarios or lightweight devices. There are many previous works investigating how to transfer knowledge from complex teacher model to student model. KD\cite{hinton2015distilling} firstly proposes a response-based knowledge distillation method,which utilizes the final fully connected layer output of the teacher model classifier as the target knowledge. DKD\cite{DKD} divides the logits information into target and non-target classes, and verifies that the information provided by non-target classes is the key to the effectiveness of the response-based knowledge distillation. BAN\cite{BAN} reveals that the predicted value of the target class in the teacher logits is equivalent to the weighting factor of each sample. The subsequent relationship-based distillation methods of FSP\cite{FSP} by constructing the relationship matrices between teacher and student feature layers separately, and RKD\cite{RKD} by constructing the relationship matrices between the features of each sample of teachers and students within the same batch separately, and both calculating the difference loss of the relationship matrices for the structured knowledge transfer method, can effectively transfer the high-level attribute knowledge even in the case of different output dimensions between the teacher and student. In the feature-based approach, FitNets\cite{romero2014fitnets} treats the output features of the middle layer of the teacher model as "Hints", which are lost as a result of the difference in feature activation at the corresponding positions in the teacher and student feature maps. FT\cite{FT} pre-trains the connectors of the teacher model to convey information. AB\cite{AB} uses binarization techniques to filter the raw features of the teacher and student models. AT\cite{AT} converts the features into attention values to optimize the knowledge transfer process. OFD\cite{OFD} investigates the effect of feature location, connector composition, and loss function adopted by various distillation algorithms on the loss of information. CRD\cite{CRD} uses contrastive learning for knowledge distillation.

However, most existing distillation methods are tailored for unimodal scenarios, operate at a single scale, and often require additional features or information, limiting their applicability to multimodal model distillation. These methods predominantly focus on the distillation of visual encoders and do not effectively extend to comprehensive multimodal distillation tasks. Relying solely on a single scale prevents the student model from fully assimilating the teacher model’s knowledge. Additionally, the necessity for supplementary features and information complicates the distillation process. In the days when pre-trained multimodal large models have received much attention, it is a top priority to investigate how to distill pre-trained multimodal large models into lightweight weight models with excellent performance.

To tackle this problem, this paper investigates how to train a lightweight student model by knowledge distillation of a pre-trained multimodal large model. We propose a novel dynamic self-adaptive multiscale distillation from pre-trained multimodal large model for efficient cross-modal representation learning for the first time. Compared with previous methods, our approach harnesses three key advantages: it utilizes a multiscale strategy, employs dynamic self-adaptive distillation, and relies solely on the output features of the pre-trained multimodal large model. This enables us to train a high-performance student model quickly and efficiently, while dynamically adjusting the weights of each loss throughout the distillation process to optimize performance. Numerous experiments validate the effectiveness of our method. The performance of our distilled student model achieves state-of-the-art performance using only image-level information and outperforms previous methods that rely on region-level information.

Overall, the contributions of our work can be summarized in the following four points:

(1) To the best of our knowledge, we propose, for the first time, a novel dynamic self-adaptive multiscale knowledge distillation from pre-trained multimodal large model for efficient cross-modal representation learning strategy, which can efficiently and rapidly distills lightweight student model with excellent performance using only the output features of teacher model and original image-level information. This strategy simplifies the distillation process, making it suitable for a wide range of scenarios.

(2) Our proposed multiscale distillation framework combines several different distillation methods, incorporating contrastive distillation, feature distillation, similarity distillation and hard negative samples distillation. The multiscale aspect of our framework ensures comprehensive learning across different dimensions of the teacher's knowledge. On the whole, contrastive distillation ensures that the student model learns the excellent structural feature space of the teacher model. In details, feature distillation and similarity distillation ensure the consistency of student features with teacher features. And hard negative samples distillation strengthens the distinction between positive and hard negative samples in the distillation process.

(3) We propose a dynamic self-adaptive distillation loss balancer, a novel component that dynamically balances multiscale distillation losses during the distillation process. This dynamic self-adaptive mechanism eliminates the need for manual loss weight adjustments, offering a more streamlined and effective optimization of the distillation process.

(4) Numerous experimental results demonstrate the effectiveness of our approach, and our distilled student model achieves state-of-the-art performance on cross-modal retrieval tasks using only image-level information, outperforming previous approaches based on region-level information. This achievement underscores the practicality and innovativeness of our distillation strategy in leveraging multimodal large models for lightweight applications.

\vspace{-4pt}
\section{Related Works}

Related prior work includes knowledge distillation, contrastive learning, and multimodal modeling.

\subsection{Knowledge Distillation}

\begin{figure*}[h]
  \centering
  \includegraphics[width=\linewidth]{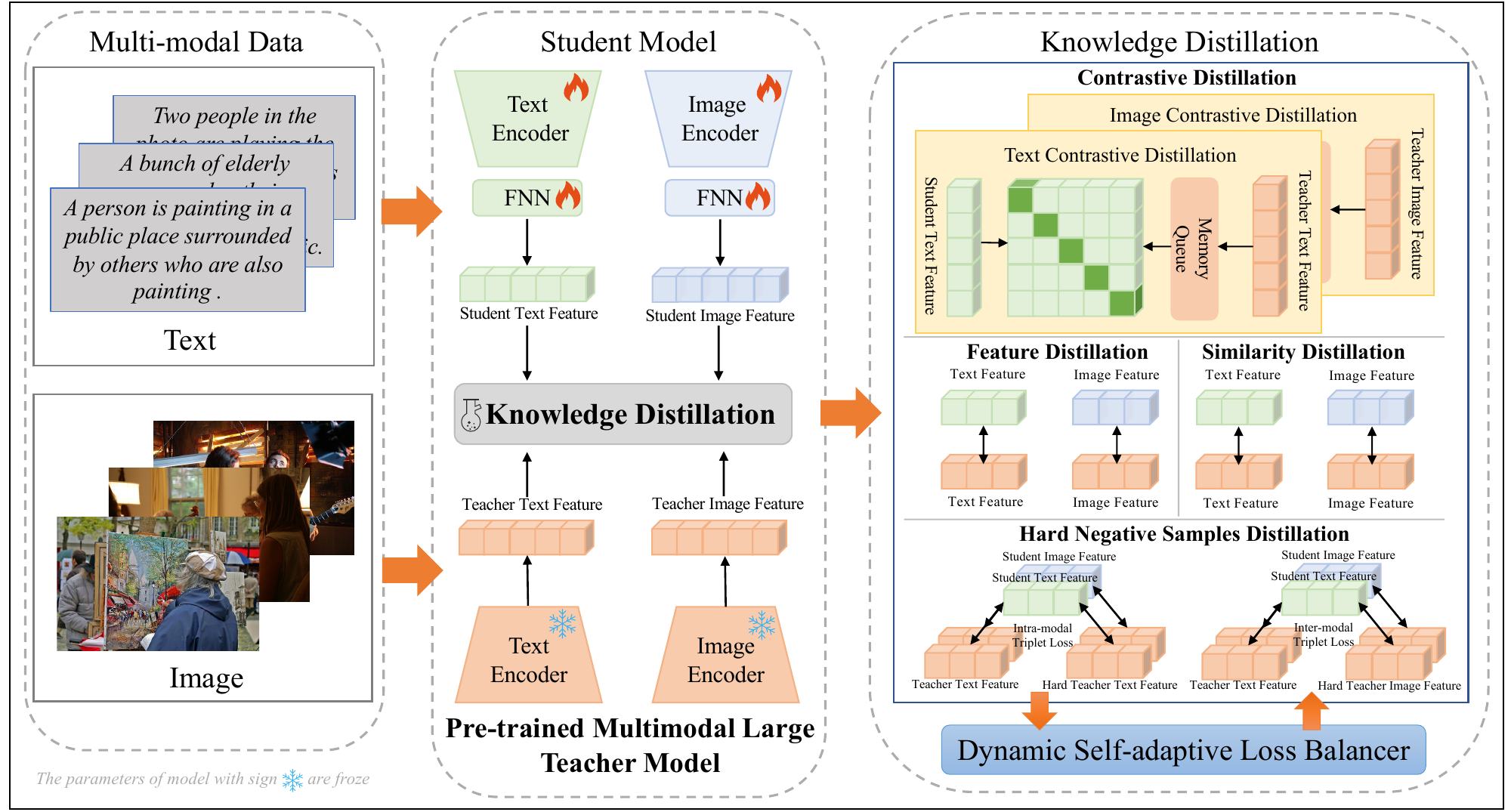}
  \caption{Our framework efficiently leverages pre-trained multimodal large models, utilizing only their output features and raw image-level data. It dynamically and adaptively accomplishes knowledge distillation across multiple scales—including contrastive distillation, feature distillation, similarity distillation, and hard negative sample distillation. This streamlined approach enables the student model to effectively master the intricate structural feature space of the teacher model.}
  \Description{framework}
\end{figure*}

The successful application of deep learning models across various domains has highlighted the challenge of deploying these large models in resource-constrained environments. Knowledge distillation emerges as a pivotal technique for this, efficiently compressing models by transferring knowledge from a complex teacher model to a simpler, lightweight student model, thus preserving the teacher's performance within a reduced model size.

Existing knowledge distillation methods are categorized into three types: response-based, relationship-based, and feature-based. KD\cite{hinton2015distilling} employs response-based distillation using the teacher's final classifier outputs for richer soft labels. DKD\cite{DKD} categorizes logits into target and non-target classes, emphasizing the importance of non-target information. BAN\cite{BAN} uses target class logits as weighting factors for samples. In relationship-based methods, FSP\cite{FSP} and RKD\cite{RKD} construct relationship matrices between teacher and student features to capture and distill structural knowledge, enhancing understanding through differences in sample relationships and geometrical configurations within the same batch.

In feature-based distillation, FitNets\cite{romero2014fitnets} uses middle layer outputs of the teacher model as "Hints" to guide the student model, noting differences in feature activations. FT\cite{FT} enhances knowledge transfer by pre-training connectors in the teacher model. AB\cite{AB} employs a binarization technique to filter features between the teacher and student models, while AT\cite{AT} transforms features into attention values to optimize distillation. CRD\cite{CRD} utilizes contrastive learning to align corresponding features between teacher and student as positive examples and separate non-corresponding ones as negatives. OFD\cite{OFD} reviews these techniques, analyzing how feature location, connector composition, and loss functions influence the efficiency of knowledge transfer.

However, previous distillation methods are tailored for unimodal scenarios, operate at a single scale, and often require additional features or information, limiting their applicability to multimodal model distillation. These methods predominantly focus on the distillation of visual encoders and do not effectively extend to comprehensive multimodal distillation tasks. Relying solely on a single scale prevents the student model from fully assimilating the teacher model’s knowledge. Additionally, the necessity for supplementary features and information complicates the distillation process. In the days when multimodal models have become more and more eye-catching. It has become a new challenge to efficiently transfer the knowledge from these large and complex multimodal model to lightweight model. Multimodal knowledge distillation needs to preserve rich multimodal semantic information and minimize performance loss. This area is still in the exploratory stage in terms of knowledge transfer for pre-trained multimodal large models.

\vspace{-4mm}
\subsection{Contrastive learning and Multimodal Model}

In recent years, there have been a number of impressive works that perform well in semantic feature learning using contrastive learning. SimCLR\cite{SimCLR} proposes a simple framework for contrastive learning of visual representations, which requires larger batch sizes for better performance. Moco\cite{MOCO} proposes a momentum contrastive method for unsupervised learning of visual representations, which replaces the just-mentioned mnemonics with a queue. NNCLR\cite{NNCLR} proposes a nearest-neighbor contrastive learning method for visual representations that uses the nearest-neighbor method to distinguish between positive and negative examples in a queue. Paco\cite{Paco} proposes a parametric contrastive learning method that uses labels as a guide to optimize the distinction between positive and negative samples in a queue.

With the success of contrastive learning in intramodal tasks, a number of cross-modal learning tasks based on contrastive learning are gaining popularity. CLIP\cite{CLIP} and ALIGN\cite{ALIGN} are proposed multimodal pre-training models based on contrastive learning, which form a task-independent model by predicting which text matches which image. In ALBEF\cite{ALBEF}, it applies contrastive loss to align image and text features before modeling their joint representation. TCL\cite{TCL} uses intramodal and intermodal contrastive learning using knowledge distillation to guide the learning process. FACLCL\cite{FACLCL} encodes knowledge of structured events and enhances visual-language pre-training with textual event maps for contrastive learning. BEiT-3\cite{BEiT-3} enhances visual-language pre-training by treating images as foreign language modeling, masking "language" modeling of images, text and image-text pairs in a unified way, and aligning image and text features with contrastive learning.

These high-performance multimodal models often rely on huge datasets for pre-training, which poses a challenge for deployment in realistic resource-constrained scenarios. An effective approach is to compress these pre-trained multimodal large model into lightweight model through knowledge distillation, which reduces the number of model parameters while maintaining the excellent performance from pre-training as much as possible. To address this challenge, in this paper, we propose a novel dynamic self-adaptive multiscale knowledge distillation from pre-trained multimodal large model for efficient cross-modal representation learning strategy.

\section{Dynamic Self-adaptive Multiscale Distillation Method}

\subsection{Teacher Model}

A series of pre-trained multimodal models with excellent performance, such as CLIP\cite{CLIP} and BEiT-3\cite{BEiT-3}, have been trained using massive Internet data, and have demonstrated outstanding performance in several downstream tasks in the vision-language domain, including visual question answering, cross-modal retrieval, and so on. In this study, we choose BEiT-3 as the teacher model T for our knowledge distillation framework, BEiT-3 is a generalized multimodal base model that achieves state-of-the-art migration performance on visual and vision-language tasks through a unified masked data modeling approach. The model uses Multiway Transformers that is capable of handling data in multiple modalities such as image, text and image-text pairs. We denote the function of the teacher model by $f_T$. The image data is represented as $X_v=\{x_v^1,......,x_v^R\}$ and the text data as $X_t=\{x_t^1,......,x_t^R\}$, with R denoting the total number of data.
\begin{gather}
T_v = \|f_T(X_v)\|_2\\
T_t = \|f_T(X_t)\|_2
\end{gather}
where $\|\cdot\|_2$ denotes L2 normalization, where the normalization process suppresses the larger values in the features and highlights the relatively important features. The resulting mapping is a good representation of the correlation between features. We show the comparison between L2 normalized and un-normalized in Table \ref{tab:l2}. The results $T_v=\{t_v^1,......,t_v^R\}$ and $T_t=\{t_t^1,......,t_t^R\}$ obtained from the reasoning of the teacher's model are used as the teacher's knowledge for the student's model learning.

\subsection{Student Model}

In this paper, we adopt the dual encoder architecture commonly used for multimodal modeling, specifically:

Visual encoder: Because visual transformers(ViT)\cite{vit} have shown great potential in visual representation extraction, many research works\cite{vvit1, BEiT-3, ALBEF, blip} have introduced ViT or its variant forms (e.g., Deit\cite{deit}, CaiT\cite{cait}, and Swin Transformer\cite{swim_transformer}) into vision-language model(VLM). we have chosen (ViT) as the visual encoder, since in this paper we focus on the the impact of model compression on performance, so we focus on the original ViT architecture.

Text Encoder: The most widely used text encoder in VLM is BERT\cite{bert}, which generates an input text sequence by first segmenting the input sentence into a sequence of individual sub-words and then inserting two special tokens at the beginning and end of the sentence. So we adopted BERT as text encoder for the student model.

The visual encoder function of the student model is denoted by $f_S^v$, while the text encoder function is represented by $f_S^t$. $S_v=\{s_v^1,......,s_v^R\}$ and $S_t=\{s_t^1,......,s_t^R\}$ are used to denote the output features of the student model, respectively.

\begin{gather}
S_{v}=\|ffn(f_{S}^{v}(X_{v}))\|_{2}\\
S_{t}=\|ffn(f_{S}^{t}(X_{t}))\|_{2}
\end{gather}
where $ffn(\cdot)$ denotes the fully-connected layer used to align the feature dimensions of the student model and the teacher model.

During the experiments, on the different settings for the student model, we observed two phenomena of interest:

1. The effect of consistency of the underlying architecture of the encoders on the distillation effect: when different underlying architectures are used for the visual and text encoders of the student model (e.g., CNN\cite{cnn} architecture for the visual encoder and Transformer\cite{transformer} architecture for the text encoder), the distillation effect is usually poorer compared to the combination of encoders with the same underlying architecture.

2. The relationship between the number of encoder parameters and distillation effectiveness: encoders with smaller number of parameters (e.g., using DeiT\cite{deit} as the visual encoder and ALBERT\cite{albert} as the text encoder) usually perform less well during distillation than the ViT and BERT models with larger number of parameters. This finding is in line with ESKD's\cite{ESKD} conclusion "Knowledge distillation is not successful when student's abilities are too low to successfully imitate the teacher." Consistent.

\vspace{-2pt}
\subsection{Multiscale Distillation Framework}

Knowledge distillation, as an effective technique for model compression and performance enhancement, has evolved in various forms. Initially, the knowledge distillation technique based on model output response proposed by Hinton et al\cite{hinton2015distilling}. opened the research in this field. With the passage of time, distillation methods based on model internal relations and features have been proposed successively, enriching the technical route of knowledge distillation. Currently, the State-of-the-Art distillation framework focuses on feature-based distillation methods, which are capable of transferring the knowledge from the teacher model to the student model in a more comprehensive way, including explicit knowledge and implicit level details. Our methodology similarly adopts a feature-based distillation paradigm, harnessing the full spectrum of knowledge encapsulated within the teacher model.

In this paper, the starting point for the distillation process is the teacher knowledge $T_v$ and $T_t$, and the aim of the framework is to train the encoder of the student model to accurately simulate these features in order to produce outputs similar to those of the teacher model. To achieve this, the distillation framework scrutinizes feature details and captures overarching patterns. To enhance the process, we introduce multiple distillation losses across different scales.

\vspace{-2pt}
\subsubsection{Contrastive Distillation}

Good teachers usually construct a well-structured feature space. Contrastive distillation allows students to imitate better-structured semantic relations from their teachers, thus improving the quality of feature representations. Contrastive learning requires supporting the learning process by constructing a large and stable dictionary. However, overstretching the dictionary poses computational performance challenges. To balance this contradiction, we design a specific queue $Q$ to store teacher knowledge $T_v$ and $T_t$, respectively. we use teacher knowledge and student knowledge matches corresponding to the same data training as positive sample pairs $D_{v}=\{s_{v}^{k},t_{v}^{k}\}_{k=1}^{|D_{v}|}$ and $D_{t}=\{s_{t}^{k},t_{t}^{k}\}_{k=1}^{|D_{t}|}$, taking the remaining mismatches in the queue as negative sample pairs. On this basis, we employ the infoNCE\cite{INFONCE} loss function to guide the contrastive distillation.

\begin{equation}\mathcal{L}_{CKD-I}=-\mathrm{log}\frac{\exp{(\mathrm{s}_{\nu}^{k}\cdot\mathrm{t}_{\nu}^{k}/\tau)}}{\sum_{i=1}^{|Q|}\exp{(\mathrm{s}_{\nu}^{k}\cdot\mathrm{t}_{\nu}^{i}/\tau)}}\end{equation}
\begin{equation}\mathcal{L}_{CKD-T}=-\log\frac{\exp{(\mathrm{s}_t^k\cdot\mathrm{t}_t^k/\tau)}}{\sum_{i=1}^{|Q|}\exp{(\mathrm{s}_t^k\cdot\mathrm{t}_t^i/\tau)}}\end{equation}
where $\tau$ denotes the temperature coefficient. ${L}_{CKD-I}$ denotes contrastive distillation in visual modality. ${L}_{CKD-T}$ denotes textual modality. Final contrastive distillation loss is:

\begin{equation}\mathcal{L}_{CD}=\mathcal{L}_{CKD-I}+\mathcal{L}_{CKD-T}\end{equation}

\subsubsection{Feature Distillation}

In order to capture the nuances of the output features, we employed the L1 distance for the positive sample pairs to assess the distance between the output features of the student model and the teacher's knowledge. 

\begin{equation}\mathcal{L}_{FD}=\sum_{i=1}^{|D_{v}|}\|\mathrm{s}_{\nu}^{i}-\mathrm{t}_{\nu}^{i}\|+\sum_{i=1}^{|D_{t}|}\|\mathrm{s}_{t}^{i}-\mathrm{t}_{t}^{i}\|\end{equation}

\subsubsection{Similarity Distillation}

In order to make the output student knowledge as similar as possible to the teacher knowledge in terms of details, we used cosine similarity for evaluation, which helped us to assess the similarity of the model output from different perspectives. 

\begin{equation}\mathcal{L}_{SD}=\sum_{i=1}^{|D_{v}|}cos(s_{v}^{i},t_{v}^{i})+\sum_{i=1}^{|D_{t}|}cos(s_{t}^{i},t_{t}^{i})\end{equation}
where $cos(\cdot)$ denotes cosine similarity.

\subsubsection{Hard Negative Sample Distillation}

Additionally, to proficiently extract challenging negative samples that are harder to distinguish, we implemented a triplet loss function that considers anchor points, positive samples, and particularly hard negative samples. This utilization of triplet loss enriches the framework's discernment and assimilation of these hard negative samples through what we term as Hinge-Based Bidirectional Triplet Loss.

\begin{gather}\mathcal{L}_{HND-I}=max\{\alpha-cos(s_{\nu}^{k},t_{\nu}^{k})+cos(s_{\nu}^{k},t_{\nu}^{h}),0\} \\
\mathcal{L}_{HND-T}=max\{\alpha-cos\bigl(s_{t}^{k},t_{t}^{k}\bigr)+cos\bigl(s_{t}^{k},t_{t}^{h}\bigr),0\} \\
\mathcal{L}_{HND-I2T}=max\{\alpha-cos\bigl(s_{v}^{k},t_{t}^{k}\bigr)+cos\bigl(s_{v}^{k},t_{t}^{h}\bigr),0\} \\
\mathcal{L}_{HND-T2I}=max\{\alpha-cos\bigl(s_{t}^{k},t_{\nu}^{k}\bigr)+cos\bigl(s_{t}^{k},t_{\nu}^{h}\bigr),0\} 
\end{gather}
where $max\{\cdot\}$ returns the maximum value,  $\alpha$ is the margin coefficient, and $t^h$ is the most indistinguishable negative sample in the batch, i.e., the negative sample with the highest similarity. ${L}_{HND-I}$ denotes the loss for the visual modality, ${L}_{HND-T}$ represents the loss for the textual modality, ${L}_{HND-I2T}$ corresponds to the loss from image to text retrieval, and ${L}_{HND-T2I}$ signifies the loss from text to image retrieval. The final hard negative distillation loss is: 

\begin{equation}\mathcal{L}_{HND}=\mathcal{L}_{HND-I}+\mathcal{L}_{HND-T}+\mathcal{L}_{HND-I2T}+\mathcal{L}_{HND-T2I}\end{equation}

\subsection{Dynamic Self-adaptive Distillation Loss Balancer}

In previous distillation work, the weights of individual losses are mostly set manually, which requires human experience and does not release the framework's capabilities well. To solve this problem and better utilize all distillation losses, we propose a dynamic self-adaptive distillation loss balancer, which aims to allow the framework to optimize each distillation loss in a balanced way. First, to avoid dominating the task with a large loss magnitude, we scale all losses to the same magnitude. Scaled loss of task m is:

\begin{equation}\mathcal{L}_{m}=log\big(max(\mathcal{L}_{i})\big)*\mathcal{L}_{m}\end{equation}

Second, inspired by DWA\cite{DWA}, we introduce the learning rate $w$,which measures the learning rate of each distillation target by loss change.

\begin{equation}w_m(t)=\frac{\mathcal{L}_m(t)}{\mathcal{L}_m(t-1)}\end{equation}
where $L_m(t)$ represents the loss of task $m$ at moment t, and $w_m(t)$ represents the learning rate of task $m$ at moment t. The weight $\lambda_{m}$ of the final task $m$ is defined as follows:

\begin{equation}\lambda_{m}(t)=\frac{Kexp(w_{m}(t)/T)}{\sum_{i}\exp(w_{i}(t)/T)}\end{equation}
where $T$ is the balancer temperature coefficient and $K$ is the sum of the weights for each task.

\subsection{Objective Function}

The final dynamic distillation objective function is as follows:

\begin{equation}\mathcal{L}(t)=\sum\lambda_{m}(t)\mathcal{L}_{m}(t)\end{equation}

We visualize the teacher feature space and student feature space. Through experiment section Figure \ref{fig:framework-visualizations} we observe that similar features cluster closely together, while dissimilar features are distinctly separated from each other. The results obtained from the inference of the student model trained by our distillation framework match this point exactly and perfectly, proving that our trained student model can mimic the feature space of the teacher model well.

\begin{table*}[]
\centering
\caption{Cross-modal Retrieval Performance on Flickr30k and MSCOCO.}
\label{table1}
\resizebox{\textwidth}{!}{%
\begin{tabular}{@{}lllllllllllllllll@{}}
\toprule
\multirow{3}{*}{Method} &
  \multirow{3}{*}{Backbone} &
  \multirow{3}{*}{Visual Embed} &
  \multicolumn{7}{c}{Flickr30k} &
  \multicolumn{7}{c}{MSCOCO 5K} \\ \cmidrule(l){4-17} 
 &
   &
   &
  \multicolumn{3}{c}{Image to text} &
  \multicolumn{3}{c}{Text to image} &
  \multirow{2}{*}{RSUM} &
  \multicolumn{3}{c}{Image to text} &
  \multicolumn{3}{c}{Text to image} &
  \multirow{2}{*}{RSUM} \\
                  &                 &        & R@1  & R@5  & R@10 & R@1  & R@5  & R@10 &       & R@1  & R@5  & R@10 & R@1  & R@5  & R@10 &   \\ \midrule
SCAN*(ECCV18)\cite{SCAN_ECCV18}   & BUTD,GRU        & Region & 67.4 & 90.3 & 95.8 & 48.6 & 77.7 & 85.2 & 465.0 & 50.4 & 82.2 & 90.0 & 38.6 & 69.3 & 80.4 & 410.9  \\
VSRN*(ICCV19)\cite{VSRN_ICCV19}   & BUTD,GRU        & Region & 71.3 & 90.6 & 96.0 & 54.7 & 81.8 & 88.2 & 482.6 & 53.0 & 81.1 & 89.4 & 40.5 & 70.6 & 81.1 & 415.7  \\
BFAN*(MM19)\cite{BFAN_MM19}     & BUTD,GRU        & Region & 68.1 & 91.4 & -    & 50.8 & 78.4 & -    & -     &      &      & -    &      &      & -    & - \\
IMRAM(CVPR20)\cite{IMRAM_CVPR20}   & BUTD,GRU        & Region & 74.1 & 93.0 & 96.6 & 53.9 & 79.4 & 87.2 & 484.2 & 53.7 & 83.2 & 91.0 & 39.6 & 69.1 & 79.8 & 416.4  \\
GSMN*(CVPR20)\cite{GSMN_CVPR20}   & Faster RCNN,GRU & Region & 76.4 & 94.3 & 97.3 & 57.4 & 82.3 & 89.0 & 496.7 & -    & -    & -    & -    & -    & -    & -  \\
MMCA(CVPR20)\cite{MMCA_CVPR20}    & BUTD,BERT       & Region & 74.2 & 92.8 & 96.4 & 54.8 & 81.4 & 87.8 & 487.4 & 54.0 & 82.5 & 90.7 & 38.7 & 69.7 & 80.8 & 416.4  \\
CAAN(CVPR20)\cite{CAAN_CVPR20}    & BUTD,GRU        & Region & 70.1 & 91.6 & 97.2 & 52.8 & 79.0 & 87.9 & 478.6 & 52.5 & 83.3 & 90.9 & 41.2 & 70.3 & 82.9 & 421.1  \\
CAMERA*(MM20)\cite{CAMERA_MM20}   & BUTD,BERT       & Region & 78.0 & 95.1 & \underline{97.9} & 60.3 & 85.9 & 91.7 & 508.9 & 55.1 & 82.9 & 91.2 & 40.5 & 71.7 & 82.5 & 423.9  \\
ADAPT*(AAAI20)\cite{ADAPT_AAAI20}  & BUTD,GRU        & Region & 76.6 & 95.4 & 97.6 & 60.7 & 86.6 & 92.0 & 508.9 &      &      &      &      &      &      & -  \\
SGRAF(AAAI21)\cite{SGRAF_AAAI21}   & BUTD,GRU        & Region & 77.8 & 94.1 & 97.4 & 58.5 & 83.0 & 88.8 & 499.6 & 57.8 & -    & 91.6 & 41.9 & -    & 81.3 & -  \\
WCGL(ICCV21)\cite{WCGL_ICCV2021}    & BUTD,GRU        & Region & 74.8 & 93.3 & 96.8 & 54.8 & 80.6 & 87.5 & 487.8 & -    & -    & -    & -    & -    & -    & -  \\
GPO(CVPR21)\cite{GPO_CVPR21}     & BUTD,BERT       & Region & 81.7 & 95.4 & 97.6 & 61.4 & 85.9 & 91.5 & 513.5 & 58.3 & 85.3 & \underline{92.3} & 42.4 & 72.7 & \underline{83.2} & 434.2  \\
DIME*(SIGIR21)\cite{DIME_SIGIR21}  & BUTD,BERT       & Region & 81.0 & 95.9 & \textbf{98.4} & \underline{63.6} & \underline{88.1} & \underline{93.0} & \underline{520.0} & 59.3 & 85.4 & 91.9 & 43.1 & 73.0 & 83.1 & 435.8  \\
SHAN*(IJCAI21)\cite{SHAN_IJCAI21}    & BUTD,GRU        & Region & 74.6 & 93.5 & 96.9 & 55.3 & 81.3 & 88.4 & 490.0 & -    & -    & -    & -    & -    & -    & -  \\
VSRN++(TPAMI22)\cite{VSRN++_TPAMI22} & BUTD,BERT       & Region & 79.2 & \textbf{96.4} & 97.5 & 60.6 & 85.6 & 91.4 & 510.7 & 54.7 & 82.9 & 90.9 & 42.0 & 72.2 & 82.7 & 425.4  \\
RCAR*(TIP23)\cite{RCAR_TIP23}    & BUTD,GRU        & Region & \textbf{82.3} & \underline{96.0} & \textbf{98.4} & 62.6 & 85.8 & 91.1 & 516.2 & \underline{61.3} & \textbf{86.1} & \textbf{92.6} & \underline{44.3} & \underline{73.2} & \underline{83.2} & \underline{440.7}  \\ \midrule
Ours              & ViT,BERT        & Image  & \underline{82.0} & 95.5 & 97.7 & \textbf{68.4} & \textbf{90.8} & \textbf{94.4} & \textbf{528.8} & \textbf{62.1} & \underline{85.9} & 92.0 & \textbf{48.0} & \textbf{75.6} & \textbf{84.5} & \textbf{448.1}  \\ \bottomrule
\end{tabular}%
}
\end{table*}

\section{Experiment and Analysis}

\subsection{Experimental settings}

\subsubsection{Downstream Task}

In terms of disambiguation and analysis, we focus on cross-modal retrieval and evaluate image-text retrieval (TR) and text-image retrieval (IR) on Flickr30K and MSCOCO.

\subsubsection{Dataset}

We conducted experiments on two widely used datasets, Flickr30k and MSCOCO, to evaluate our approach and the latest advanced methods.

Flickr30k\cite{Flickr30k}: It contains 31,783 images from the Flickr website, each described by five different sentences. Following the setup in BEiT-3\cite{BEiT-3}, this dataset is divided into 29,783 training images, 1,000 validation images and 1,000 test images.

MSCOCO\cite{MSCOCO}: This dataset consists of 123,287 images, each associated with five annotated sentences. Again, we split dataset like BEiT-3\cite{BEiT-3}, i.e., 113,287 images for training, 5,000 images for validation, and 5,000 images for testing. Similarly, the evaluation setup considered in this paper: the MSCOCO 5k and the evaluation results are computed from all 5k test images.

\subsubsection{Evaluation indicators}

We adopt Recall at K (R@K) (K = 1, 5, and 10) as an evaluation metric, which are commonly used in cross-modal retrieval task. R@K is defined as the percentage of basic facts retrieved in the first K results. Higher R@K indicates better performance. We also use RSUM (sum of R@K) as an evaluation metric to compute the total value of R@K for cross-modal retrieval. RSUM provides a general view of the overall retrieval performance. As with R@K, higher RSUM indicates better performance.

\subsubsection{Implementation details}

Our experiments were all performed on the 1 NVIDIA 4090 GPU, and we trained the model using the AdamW\cite{Adamw} optimizer with a weight decay of 1e-4. The learning rate was preset to 1e-4 and decayed linearly at a rate of 0.1. The visual encoder is initialized by Vit pre-trained on ImageNet and the text encoder is by BERT pre-trained. During training, we crop images with a resolution of 224 × 224. The model is trained with batch size 64. The queue size hyperparameter $|Q$ is 8192, balancer temperature coefficient $T$ is 1, contrastive temperature coefficient $\tau$ is 0.05 and margin coefficient $\alpha$ is 0. We trained a total of 20 epochs and performed a linear decay at the 10th epoch.

\subsection{Experimental Results}

\subsubsection{Baselines}

To demonstrate the effectiveness of our approach, we compare our student model with 16 recent state-of-the-art methods in cross-modal retrieval task on two widely used datasets by R@K. The comparison results are summarized in Table \ref{table1}. The best performance is highlighted in bold, and the best performance of previous methods is emphasized with underlines. The state-of-the-art methods include SCAN\cite{SCAN_ECCV18}, VSRN\cite{VSRN_ICCV19}, BFAN\cite{BFAN_MM19}, IMRAM\cite{IMRAM_CVPR20}, GSMN\cite{GSMN_CVPR20}, MMCA\cite{MMCA_CVPR20}, CAAN\cite{CAAN_CVPR20}, CAMERA\cite{CAMERA_MM20}, ADAPT\cite{ADAPT_AAAI20}, SGRAF\cite{SGRAF_AAAI21}, WCGL\cite{WCGL_ICCV2021}, GPO\cite{GPO_CVPR21}, DIME\cite{DIME_SIGIR21}, SHAN\cite{SHAN_IJCAI21}, VSRN++\cite{VSRN++_TPAMI22}, RCAR\cite{RCAR_TIP23}. All of these methods are based on region-level information, and it is worth mentioning that our method achieves the state-of-the-sota performance using only raw image-level information.\\
* Denotes Ensemble Models.\\“BUTD” represents the Bottom-Up and Top-Down attention model\cite{BUTD} for image encoding. This model builds on Faster-RCNN\cite{faster——rcnn} and pre-trains on Visual Genome\cite{visual_genome}.\\“GRU” represents the Gate Recurrent Unit\cite{GRU} for text encoding.\\“BERT” represents the BERT\cite{bert} model for text encoding.

We also compared our distillation method with other distillation methods\cite{hinton2015distilling, romero2014fitnets} or student models\cite{CLIP-KD, DCD} derived from distillation in Table \ref{table:dis}.

\subsubsection{Comparison with State-of-the-art Methods}

Our method shows significant improvements on Flickr30k dataset for Text-Image retrieval, with an increase of 7.5$\%$ in R@1, 3.1$\%$ in R@5, and 3.2$\%$ in R@10, along with a 1.7$\%$ rise in RSUM. On the MSCOCO dataset for Image-Text retrieval, we observed a 1.3$\%$ increase in R@1. For Text-Image retrieval, there's an impressive boost of 8.4$\%$ in R@1, 3.3$\%$ in R@5, and 1.6$\%$ in R@10, with RSUM also improving by 1.7$\%$.

In table \ref{table1}, all the methods we compared against utilize region-level information without exception, while our approach only requires original image-level information to achieve superior performance simply and efficiently. The results are particularly impressive considering that our method does not rely on region-level information, in contrast to the other listed approaches. Instead, by leveraging only the original image-level information, our dynamic self-adaptive multiscale distillation strategy proves to be both simple and effective. This highlights the efficiency of our approach and its potential applicability to resource-constrained scenarios, allowing users to benefit from high-level multimodal representations without the overhead of more complex models. It also underscores the capacity of our model to dynamically adapt and distill the teacher model's structural feature information across multiple scales, ensuring the student model captures the essence of the data representation effectively.

\begin{table}[h]
\caption{Comparison with other Distillation Methods}
\label{table:dis}
\resizebox{\columnwidth}{!}{%
\begin{tabular}{@{}llllllll@{}}
\toprule
\multirow{3}{*}{Method} & \multicolumn{7}{c}{Flickr30k}                                                                 \\ \cmidrule(l){2-8} 
                       & \multicolumn{3}{c}{Image to text} & \multicolumn{3}{c}{Text to image} & \multirow{2}{*}{RSUM} \\
                       & R@1       & R@5       & R@10      & R@1       & R@5       & R@10      &                       \\ \midrule
CLIP-KD\cite{CLIP-KD}                & 81.7      & 95.4      & 97.6      & 61.4      & 85.9      & 91.5      & 513.5                 \\
KD\cite{hinton2015distilling}                     & 79.9      & 94.9      & 96.5      & 37.0      & 62.8      & 73.3      & 444.4                 \\
FSP\cite{romero2014fitnets}                    & 76.6      & 93.1      & 96.0      & 66.4      & 88.3      & 93.4      & 513.8                 \\
DCD\cite{DCD}                    & 75.6      & 91.0      & 94.6      & 53.7      & 81.4      & 88.2      & 484.5                 \\ \midrule
Ours                   & \textbf{82.0}      & \textbf{95.5}      & \textbf{97.7}      & \textbf{68.4}      & \textbf{90.8}      & \textbf{94.4}      & \textbf{528.8}                 \\ \bottomrule
\end{tabular}%
}
\end{table}

Table \ref{table:dis} shows our method emerges as the clear victor across all metrics, conclusively demonstrating its ability to enable the student model to more effectively master the teacher model's superior feature space as compared to other distillation methods.

\begin{table}[h]
\caption{Comparison with Teacher Model}
\label{tab:tea}
\resizebox{\columnwidth}{!}{%
\begin{tabular}{@{}llllllllll@{}}
\toprule
\multirow{3}{*}{Model} &
  \multirow{3}{*}{Params} &
  \multirow{3}{*}{Inference time} &
  \multicolumn{7}{c}{Flickr30k} \\ \cmidrule(l){4-10} 
 &
   &
   &
  \multicolumn{3}{c}{Image to text} &
  \multicolumn{3}{c}{Text to image} &
  \multirow{2}{*}{RSUM} \\
      &  &  & R@1  & R@5  & R@10 & R@1  & R@5  & R@10 &       \\ \midrule
BEiT-3\cite{BEiT-3} & $\sim$1900M & 0.0165s  & 98.0 & 100.0 & 100.0 & 90.3 & 98.7 & 99.5 & 586.5 \\ \midrule
Ours  & $\sim$197M  & 0.0021s  & 82.0 & 95.5 & 97.7 & 68.4 & 90.8 & 94.4 & 528.8 \\ \bottomrule
\end{tabular}%
}
\end{table}

In the comparison outlined in Table \ref{tab:tea}, our student model, utilizing merely 10$\%$ of the parameters of the BEiT-3 teacher model, achieves impressive performance, attaining 90$\%$ of the teacher's efficacy in the cross-modal retrieval task on the Flickr30k dataset. Furthermore, the student model exhibits a notable efficiency in inference times, significantly reducing the latency to just over 12$\%$ of the teacher model's time. This comparative analysis showcases our model's efficacy in distilling the teacher model's complex feature space into a more compact and efficient student model without significantly compromising on performance.

\subsubsection{Ablation Experiment}

Table \ref{tab:variants} presents ablation studys of our proposed method, showcasing the impact of various components on the recall performance on Flickr30k dataset. We analyzed Contrastive Distillation (CD), Feature Distillation (FD), Similarity Distillation (SD), and Hard Negative Distillation (HND).

\begin{table}[h]
\caption{Recall performance of different variants}
\label{tab:variants}
\resizebox{\columnwidth}{!}{%
\begin{tabular}{@{}lllllllllll@{}}
\toprule
\multirow{3}{*}{CD} &
  \multirow{3}{*}{FD} &
  \multirow{3}{*}{FSD} &
  \multirow{3}{*}{HND} &
  \multicolumn{7}{c}{Flickr30k} \\ \cmidrule(l){5-11} 
 &
   &
   &
   &
  \multicolumn{3}{c}{Image to text} &
  \multicolumn{3}{c}{Text to image} &
  \multirow{2}{*}{RSUM} \\
  &   &   &   & R@1  & R@5  & R@10 & R@1  & R@5  & R@10 &       \\ \midrule
\ding{55} & \ding{51} & \ding{51} & \ding{51} & 80.1 & 94.2 & 97.1 & \textbf{68.9} & 89.8 & 94.1 & 524.2      \\
\ding{51} & \ding{55} & \ding{51} & \ding{51} & 72.6 & 91.7 & 95.4 & 58.8 & 84.2 & 90.2 & 492.9      \\
\ding{51} & \ding{51} & \ding{55} & \ding{51} & 80.6 & 94.2 & 96.8 & 68.2 & 89.5 & 94.0 & 523.3      \\
\ding{51} & \ding{51} & \ding{51} & \ding{55} & \textbf{83.1} & 95.3 & 97.1 & 65.8 & 86.7 & 91.3 & 519.3      \\
\ding{55} & \ding{55} & \ding{51} & \ding{51} & 80.8 & 94.3 & 96.8 & \textbf{68.9} & 90.0 & 94.0 & 524.8      \\
\ding{55} & \ding{51} & \ding{55} & \ding{51} & 81.1 & 95.2 & 97.0 & 67.8 & 89.3 & 94.1 & 524.5      \\
\ding{55} & \ding{51} & \ding{51} & \ding{55} & 79.1 & 93.2 & 96.4 & 67.5 & 89.0 & 93.5 & 518.7      \\
\ding{51} & \ding{55} & \ding{55} & \ding{51} & 81.3 & \textbf{95.5} & \textbf{97.8} & 64.6 & 86.2 & 91.4 & 516.8      \\
\ding{51} & \ding{55} & \ding{51} & \ding{55} & 80.7 & 94.2 & 96.3 & 68.1 & 89.4 & 93.6 & 522.3      \\
\ding{51} & \ding{51} & \ding{55} & \ding{55} & 82.9 & 95.3 & \textbf{97.8} & 65.9 & 87.1 & 92.2 & 521.2      \\
\ding{51} & \ding{55} & \ding{55} & \ding{55} & 76.9 & 94.5 & 96.7 & 61.3 & 84.4 & 89.4 & 503.2      \\
\ding{55} & \ding{51} & \ding{55} & \ding{55} & 71.7 & 91.9 & 95.1 & 63.9 & 86.8 & 92.0 & 501.4      \\
\ding{55} & \ding{55} & \ding{51} & \ding{55} & 75.1 & 92.6 & 95.6 & 63.7 & 87.1 & 92.3 & 506.4      \\
\ding{55} & \ding{55} & \ding{55} & \ding{51} & 53.8 & 80.3 & 89.6 & 46.7 & 73.9 & 82.4 & 426.7      \\
\ding{51} & \ding{51} & \ding{51} & \ding{51} & 82.0 & \textbf{95.5} & 97.7 & 68.4 & \textbf{90.8} & \textbf{94.4} & \textbf{528.8} \\ \bottomrule
\end{tabular}%
}
\end{table}

The results indicate that each component contributes positively to the overall performance, with all components achieving the highest recall scores across the board. Specifically, the highest R@1 results consistently feature at least two of the following components: FD, SD, and HND. Additionally, the top cases for R@5 and R@10 invariably include CD. These outcomes validate the efficacy of our multiscale distillation framework.

Remarkably, when all components are integrated, the recall performance reaches its peak. This comprehensive model not only consolidates the strengths of each individual component but also 
highlights the synergy between them, culminating in a robust and effective representation learning framework.
\begin{table}[h]
\caption{Ablation Experiment of Balancer}
\label{tab:balan}
\resizebox{\columnwidth}{!}{%
\begin{tabular}{@{}llllllll@{}}
\toprule
\multirow{3}{*}{Balancer} & \multicolumn{7}{c}{Flickr30k}                                                                 \\ \cmidrule(l){2-8} 
                         & \multicolumn{3}{c}{Image to text} & \multicolumn{3}{c}{Text to image} & \multirow{2}{*}{RSUM} \\
                         & R@1       & R@5       & R@10      & R@1       & R@5       & R@10                             \\ \midrule
\ding{55}                        & 61.6      & 86.0      & 92.2      & 48.7      & 76.0      & 83.6      & 448.1                 \\
\ding{51}                        & \textbf{82.0} & \textbf{95.5} & \textbf{97.7} & \textbf{68.4} & \textbf{90.8} & \textbf{94.4} & \textbf{528.8}                 \\ \bottomrule
\end{tabular}%
}
\end{table}

Table \ref{tab:balan} details the absence of the balancer (indicated by \ding{55}) leads to lower recall rates across all metrics when compared to the full implementation with the balancer (indicated by \ding{51}). With the balancer active, there is a significant leap in performance, substantiated by the RSUM scores, which rise from 448.1 to 528.8 when the balancer is included. These results underscore the efficacy of the dynamic self-adaptive loss balancer in optimizing the distillation process.
\begin{table}[h]
\caption{Ablation Experiment of L2 Normalization}
\label{tab:l2}
\resizebox{\columnwidth}{!}{%
\begin{tabular}{@{}llllllll@{}}
\toprule
\multirow{3}{*}{L2 Norm} & \multicolumn{7}{c}{Flickr30k}                                                                 \\ \cmidrule(l){2-8} 
                         & \multicolumn{3}{c}{Image to text} & \multicolumn{3}{c}{Text to image} & \multirow{2}{*}{RSUM} \\
                         & R@1       & R@5       & R@10      & R@1       & R@5       & R@10                             \\ \midrule
\ding{55}                        & 25.0      & 59.3      & 73.7      & 31.3      & 59.4      & 70.2      & 318.9                 \\
\ding{51}                        & \textbf{82.0} & \textbf{95.5} & \textbf{97.7} & \textbf{68.4} & \textbf{90.8} & \textbf{94.4} & \textbf{528.8}                 \\ \bottomrule
\end{tabular}%
}
\end{table}

Table \ref{tab:l2} shows the implementation of L2 normalization dramatically enhances cross-modal retrieval efficacy improvement in the RSUM score from 318.9 to 528.8. Confirming the indispensable contribution of L2 normalization to the optimization of representation learning within our distillation framework.

\subsubsection{Parameter sensitivity analysis}

We performed a sensitivity analysis of the hyperparameters in our experiments, including the contrastive temperature coefficient $\tau$, contrastive queue size $|Q|$, margin coefficient $\alpha$, balancer temperature coefficient $T$.
\begin{table}[H]
\caption{Sensitivity Analysis of $\tau$}
\label{tab:tau}
\resizebox{\columnwidth}{!}{%
\begin{tabular}{@{}llllllll@{}}
\toprule
\multirow{3}{*}{$\tau$} & \multicolumn{7}{c}{Flickr30k}                                                                 \\ \cmidrule(l){2-8} 
                   & \multicolumn{3}{c}{Image to text} & \multicolumn{3}{c}{Text to image} & \multirow{2}{*}{RSUM} \\
                   & R@1       & R@5       & R@10      & R@1       & R@5       & R@10                             \\ \midrule
0.05               & \textbf{82.0} & \textbf{95.5} & \textbf{97.7} & \textbf{68.4} & \textbf{90.8} & \textbf{94.4} & \textbf{528.8}                       \\
0.2                & 78.1      & 94.9      & 97.0      & 67.4      & 89.7      & 94.0      & 521.1                      \\
0.5                & 79.1      & 94.8      & 97.0      & 67.6      & 90.0      & 94.1      & 522.6                      \\
1                  & 78.0      & 94.8      & 96.9      & 67.9      & 89.9      & 94.2      & 521.7                      \\ \bottomrule
\end{tabular}%
}
\end{table}

Table \ref{tab:tau} shows the temperature coefficient $\tau$ is instrumental in training student model, where a lower $\tau$ leads to better cross-modal retrieval performance. A higher $\tau$ inversely affects recall.

\begin{table}[H]
\caption{Sensitivity Analysis of $|Q|$}
\label{tab:queue}
\resizebox{\columnwidth}{!}{%
\begin{tabular}{@{}llllllll@{}}
\toprule
\multirow{3}{*}{$|Q|$} & \multicolumn{7}{c}{Flickr30k}                                                                 \\ \cmidrule(l){2-8} 
                   & \multicolumn{3}{c}{Image to text} & \multicolumn{3}{c}{Text to image} & \multirow{2}{*}{RSUM} \\
                   & R@1       & R@5       & R@10      & R@1       & R@5       & R@10                             \\ \midrule
1024                & 80.3      & 95.5      & 97.9      & 68.1      & 89.5      & 94.1      & 525.4                      \\
2048                & 80.9      & 94.8      & 98.2      & 67.7      & 90.2      & \textbf{94.5}      & 526.3                      \\
4096                & 80.6      & \textbf{96.0}      & \textbf{98.3}      & \textbf{68.9}      & 90.2      & 94.4      & 528.4                      \\
8192                & \textbf{82.0} & 95.5 & 97.7 & 68.4 & \textbf{90.8} & 94.4 & \textbf{528.8}                       \\
16384               & 80.0      & 95.1      & 97.5      & 68.3      & 90.0      & 94.1      & 525.0                      \\ \bottomrule
\end{tabular}%
}
\end{table}

Table \ref{tab:queue} shows queue size $|Q|$ has a pronounced effect on cross-modal retrieval outcomes, with increased $|Q|$ correlating with improved performance up to an optimal point, beyond which gains plateau. 

\begin{table}[H]
\caption{Sensitivity Analysis of $\alpha$}
\label{tab:margin}
\resizebox{\columnwidth}{!}{%
\begin{tabular}{@{}llllllll@{}}
\toprule
\multirow{3}{*}{$\alpha$} & \multicolumn{7}{c}{Flickr30k}                                                                 \\ \cmidrule(l){2-8} 
                   & \multicolumn{3}{c}{Image to text} & \multicolumn{3}{c}{Text to image} & \multirow{2}{*}{RSUM} \\
                   & R@1       & R@5       & R@10      & R@1       & R@5       & R@10                             \\ \midrule
0               & \textbf{82.0} & \textbf{95.5} & \textbf{97.7} & \textbf{68.4} & \textbf{90.8} & \textbf{94.4} & \textbf{528.8}                       \\
0.05               & 75.3      & 93.8      & 97.0      & 68.2      & 90.4      & 93.9      & 518.6                      \\
0.2                & 79.5      & 94.4      & 97.5      & 67.1      & 90.0      & \textbf{94.4}      & 522.9                      \\
0.5                & 77.4      & 94.7      & 97.0      & 68.0      & 90.2      & \textbf{94.4}      & 521.7                      \\
1                  & 79.0      & 94.7      & 97.6      & 67.5      & 89.8      & 94.1      & 522.7                      \\ \bottomrule
\end{tabular}%
}
\end{table}

Table \ref{tab:margin} shows the impact of the margin coefficient $\alpha$ on the cross-modal retrieval efficacy. Optimal retrieval outcomes are achieved with $\alpha$ set to 0, while an increase in $\alpha$ leads to a decline in performance. 

\begin{table}[H]
\caption{Sensitivity Analysis of $T$}
\label{tab:balan_t}
\resizebox{\columnwidth}{!}{%
\begin{tabular}{@{}llllllll@{}}
\toprule
\multirow{3}{*}{$T$} & \multicolumn{7}{c}{Flickr30k}                                                                 \\ \cmidrule(l){2-8} 
                   & \multicolumn{3}{c}{Image to text} & \multicolumn{3}{c}{Text to image} & \multirow{2}{*}{RSUM} \\
                   & R@1       & R@5       & R@10      & R@1       & R@5       & R@10                             \\ \midrule
0.05               & 58.0      & 84.0      & 90.5      & 50.0      & 78.6      & 86.4      & 447.5                      \\
0.2                & 79.2      & 94.6      & 96.9      & 64.1      & 86.2      & 90.0      & 511.0                      \\
0.5                & 80.6      & \textbf{95.8}      & 97.5      & 68.1      & 90.2      & 94.2      & 526.4                      \\
1                  & \textbf{82.0} & 95.5 & \textbf{97.7} & \textbf{68.4} & \textbf{90.8} & \textbf{94.4} & \textbf{528.8}                       \\ \bottomrule
\end{tabular}%
}
\end{table}

Table \ref{tab:balan_t} shows increasing $T$ enhances recall metrics for cross-modal retrieval tasks, achieving optimal performance at the upper end of the temperature scale.

\subsubsection{Visualization}

\begin{figure}[h]
  \centering
  \begin{subfigure}{.5\linewidth}
    \centering
    \includegraphics[width=\linewidth]{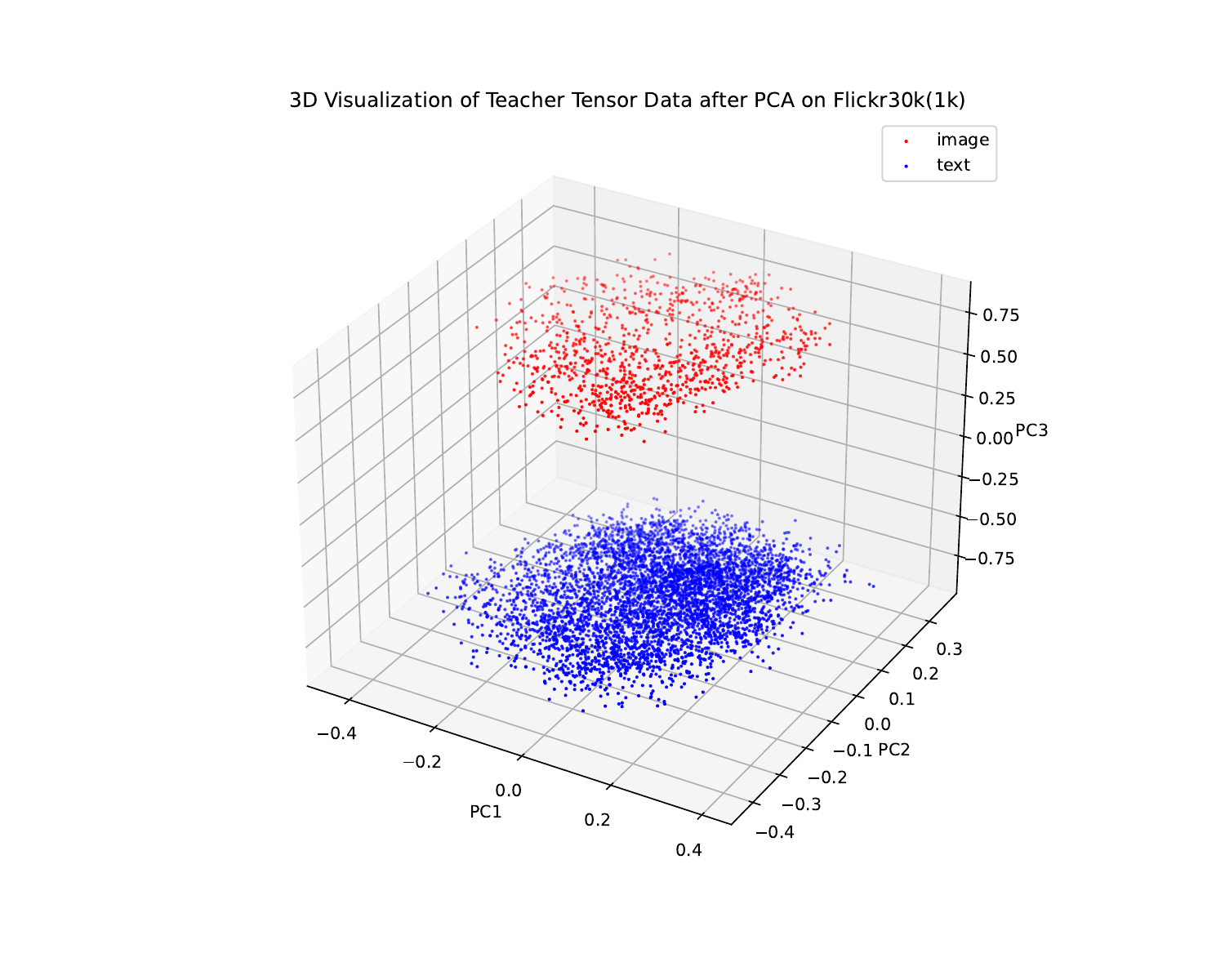}
    \caption{Teacher Feature Space}
    \label{fig:teacher}
  \end{subfigure}%
  \begin{subfigure}{.5\linewidth}
    \centering
    \includegraphics[width=\linewidth]{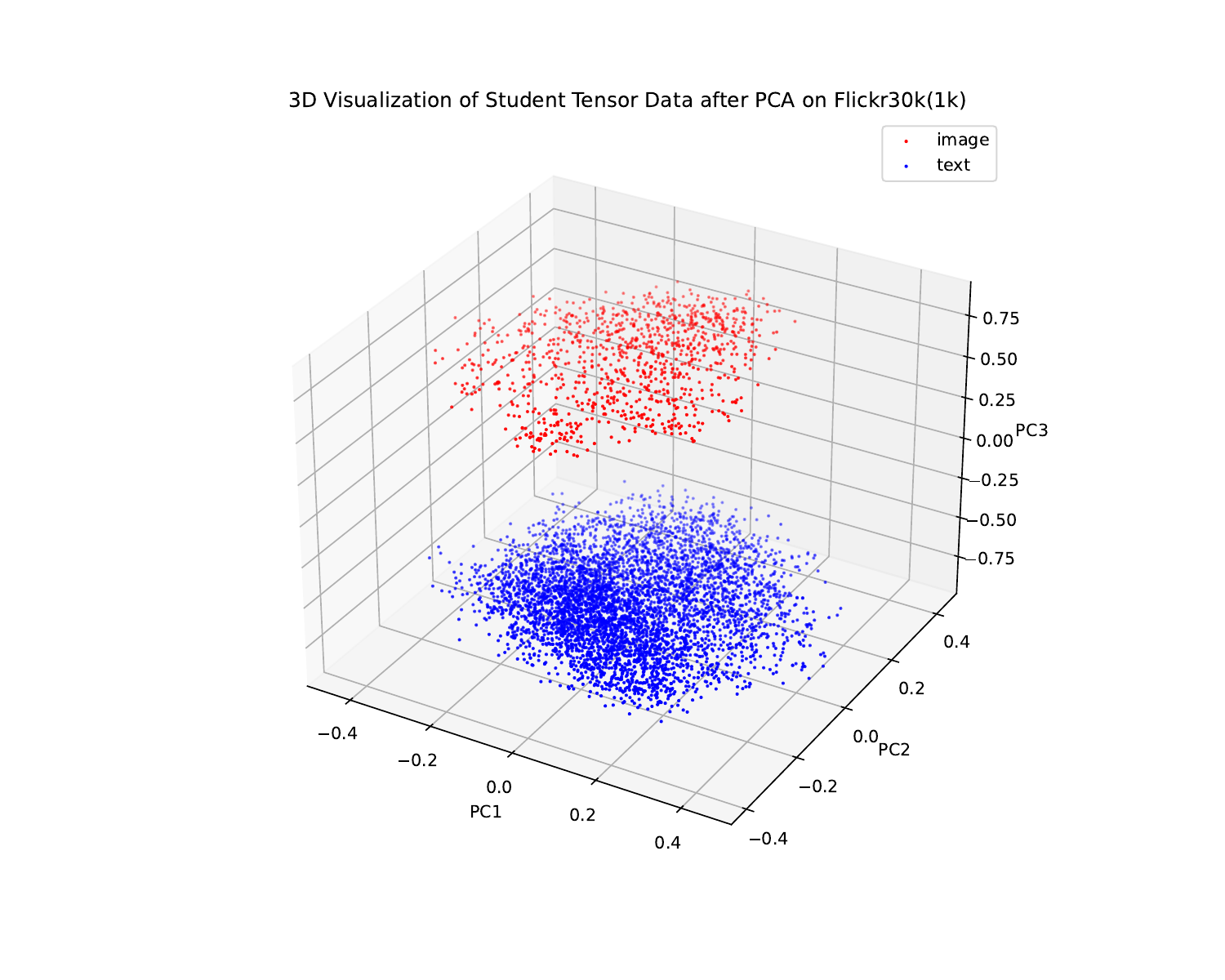}
    \caption{Student Feature Space}
    \label{fig:student}
  \end{subfigure}
  \caption{3D Visualizations of Feature Space after PCA}
  \label{fig:framework-visualizations}
\end{figure}

Figure \ref{fig:framework-visualizations} displays 3D visualizations of the feature space for both the teacher and student models after dimensionality reduction to three dimensions using PCA. Red indicates image feature, blue indicates text feature. The close alignment of the student's feature distribution with the teacher's space reflects the efficacy of the distillation process in preserving the essential structural characteristics imparted by the teacher model.

\section{Conclusion}

In this paper, we propose the dynamic self-adaptive multiscale distillation from pre-trained multimodal large model for efficient cross-modal representation learning strategy. The student model is allowed to fully learn the excellent structural feature space of the teacher model through multi-scale distillation. Our dynamic adaptive loss balancer automatically adjust the distillation loss to make the distillation process more efficient. Our framework can efficiently distill a high-performance student model using only the output features of the pre-trained multimodal large model and the original image-level information.

\bibliographystyle{ACM-Reference-Format}
\bibliography{main}

\end{document}